\pdfoutput=1

\documentclass[11pt]{article}

\usepackage[final]{acl}

\usepackage{times}
\usepackage{latexsym}

\usepackage[T1]{fontenc}

\usepackage[utf8]{inputenc}

\usepackage{microtype}

\usepackage{inconsolata}

\usepackage{graphicx}
\usepackage{amsmath,amssymb,booktabs}
\usepackage[font=small]{caption}

\title{PATCH! \{P\}sychometrics-\{A\}ssis\{T\}ed Ben\{CH\}marking of Large Language Models against Human Populations: \\A Case Study of Proficiency in 8th Grade Mathematics}

\author{
 \textbf{Qixiang Fang},
 \textbf{Daniel L Oberski},
 \textbf{Dong Nguyen},
\\
\\
Utrecht University, Netherlands,
\\
 \small{
   \textbf{Correspondence:} \href{mailto:email@domain}{q.fang@uu.nl}
 }
}

\begin{document}
\maketitle
\begin{abstract}
Many existing benchmarks of large (multimodal) language models (LLMs) focus on measuring LLMs' academic proficiency, often with also an interest in comparing model performance with human test takers'. 
While such benchmarks have proven key to the development of LLMs, they suffer from several limitations, including questionable measurement quality (e.g., Do they measure what they are supposed to in a reliable way?), lack of quality assessment on the item level (e.g., Are some items more important or difficult than others?) and unclear human population reference (e.g., To whom can the model be compared?). 
In response to these challenges, we propose leveraging knowledge from psychometrics---a field dedicated to the measurement of latent variables like academic proficiency---into LLM benchmarking. 
We make four primary contributions.
First, we reflect on current LLM benchmark developments and contrast them with psychometrics-based test development.
Second, we introduce PATCH: a novel framework for \textbf{P}sychometrics-\textbf{A}ssis\textbf{T}ed ben\textbf{CH}marking of LLMs. PATCH addresses the aforementioned limitations. In particular, PATCH enables valid comparison between LLMs and human populations.
Third, we demonstrate PATCH by measuring several LLMs' proficiency in 8th grade mathematics against 56 human populations. 
We show that adopting a psychometrics-based approach yields evaluation outcomes that diverge from those based on current benchmarking practices.
Fourth, we release 4 high-quality datasets to support measuring and comparing LLM proficiency in grade school mathematics and science with human populations.
\end{abstract}

\section{Introduction}
\begin{figure}[ht]
    \centering
    \includegraphics[width=\linewidth]{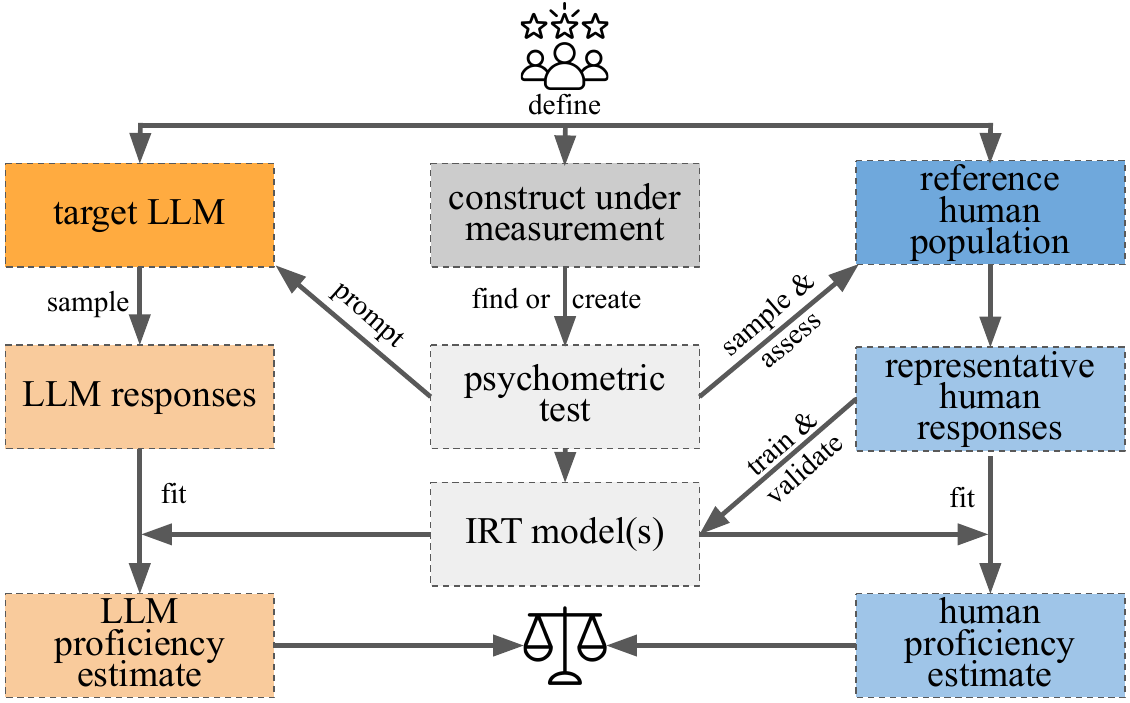}
    \caption{\textbf{PATCH: A \{P\}sychometrics-\{A\}ssis\{T\}ed framework for ben\{CH\}marking LLMs against humans.}}
    \label{fig:framework}
\end{figure}

Large language models (LLMs), including their multimodal variants like vision language models, have witnessed significant advancements in recent years. 
These models are typically evaluated on established benchmarks that assess their performance across a diverse set of tasks such as \textit{commonsense reasoning}~\cite{Zellers2019HellaSwagCA,Sakaguchi2019WinoGrande,chen2021evaluating}, \textit{coding}~\cite{chen2021evaluating,team2023gemini} and \textit{academic proficiency}. 
Academic proficiency, in particular, has become a crucial part of LLM evaluation, as evidenced by the large number of related benchmarks like MMLU, ARC, GSM8K, DROP and MATH~\cite{hendrycks2021measuring,Clark2018ThinkYH,Cobbe2021TrainingVT,Dua2019DROPAR,hendrycks2021measuring}, as well as recent model technical reports' increasing focus on them~\cite{achiam2023gpt,team2023gemini}. In these benchmarks and reports, the contrast between LLM performance and human performance is often highlighted, sparking media coverage and discussions.

Despite their success in advancing LLM research and shedding light on the artificial versus human intelligence debate, existing benchmarks have notable limitations.
The first concern is measurement quality: \textit{Do these benchmarks measure what they are supposed to in a reliable way?}
Many benchmarks are created via crowd-sourced knowledge, by asking a convenience group of individuals (e.g., crowd workers, paper authors) to create new test items (e.g., GSM8K, DROP) or collecting them from (often undocumented) sources (e.g., websites, textbooks, school exams) (e.g., MATH, MMLU, ARC). 
Without domain expert input and rigorous testing of item quality, undesirable outcomes can occur, including a mismatch between a benchmark and its claimed measurement goal, missing information in a question, wrong answer keys, and low data annotation agreement~\cite[e.g.,][]{nie-etal-2020-learn,wang2024mmlupro,HellaSwagErrors2024}.

Second, current benchmarks do not account for differences across test items, such as item discrimination and difficulty (see Section~\ref{sec:irt}). 
For example, consider three items A (easy), B (hard) and C (hard). While answering correctly to A and B would result in the same accuracy score as answering correctly to B and C, the latter (i.e., answering correctly to more difficult items) would imply higher proficiency. Furthermore, items that are too easy or too difficult (i.e., low discrimination) will fail to differentiate models (and humans) of different proficiency levels. 
Thus, without accounting for item differences, benchmarking results, especially model (versus human) rankings, can be misleading. 

Third, while many benchmarks compare LLMs against humans, the human populations under comparison remain unclear~\cite{tedeschi-etal-2023-whats}.
For instance, human performance in MATH is based on the benchmark's authors; in MMLU, crowd workers; in MATH, 6 university students. 
Using such convenience samples (with little information about sample characteristics), the resulting human performance cannot be generalised to other human samples or populations. 

To address these challenges, we propose leveraging insights from psychometrics---a field dedicated to the measurement of \textit{latent} variables like academic proficiency---into LLM benchmarking practices. 
In particular, we draw on two research areas in psychometrics: \textit{item response theory} (IRT) (Section~\ref{sec:irt}) and \textit{test development} (Section~\ref{sec:test_development}). 
The former enables more accurate estimation of academic proficiency on a standardised scale by taking into account both the characteristics of the test items as well as the abilities of the LLMs and individuals being assessed, compared to common practices in LLM benchmarks (e.g., using mean scores, percentages of correct responses). It can also provide diagnostic information about the quality of each test item. The latter, test development knowledge, can help to build high quality LLM benchmarks where valid comparison to specific human populations can be made (Section~\ref{sec:comparison_with_psychometrics}).

Our paper makes four primary contributions.
\textit{First}, we reflect on current LLM benchmark development processes and contrast them with psychometrics-based test development, thereby revealing the limitations of current LLM benchmarks and the potential benefits that PATCH/psychometrics can bring to LLM benchmarking.
\textit{Second}, we present \textbf{PATCH}: a novel framework for \textbf{P}sychometrics-\textbf{A}ssis\textbf{T}ed ben\textbf{CH}marking of LLMs (Figure~\ref{fig:framework}). PATCH is built upon IRT and test development insights from psychometrics and addresses the aforementioned limitations of existing benchmarks. 
\textit{Third}, we demonstrate the IRT part of PATCH by testing several LLMs' proficiency in 8th grade mathematics using the released test items and data from \textit{Trends in International Mathematics and Science Study}\footnote{\url{http://timssandpirls.bc.edu/timss2015/encyclopedia/}} (TIMSS) 2011. We show empirically that an IRT-based approach can lead to evaluation outcomes that diverge from those obtained through conventional benchmarking practices and that are more informative, underscoring the potential of PATCH/psychometrics to reshape the LLM benchmarking landscape.
\textit{Fourth}, we make our evaluation code based on the PATCH framework available\footnote{\url{https://github.com/fqixiang/patch_llm_benchmarking_with_psychometrics}}, along with three other mathematics and science datasets based on TIMSS 2011 and 2008\footnote{\url{https://zenodo.org/records/12531906}. See also Appendix~\ref{app:TIMSS}.}.

\section{Related Work}
\label{sec:related_work}
We are not the first to propose leveraging psychometrics for research on LLMs and other areas in NLP. For instance, psychometric scales have been used to examine the psychological profiles of LLMs such as personality traits and motivations~\cite{huang2024on,pellert2023ai,dillion2023can}.
The text in these scales can also be used to improve encoding and prediction of personality traits~\cite{kreuter_items_2022,vu_predicting_2020,yang_learning_2021,fang2023text}. 
Psychometrics-based reliability and validity tests have also been proposed or/and used to assess the quality of NLP bias measures~\cite{du2021assessing,van2024undesirable}, text embeddings~\cite{fang_evaluating_2022}, political stance detection~\cite{sen2020reliability}, annotations~\cite{amidei2020identifying}, user representations~\cite{fang2023designing}, and general social science constructs~\cite{birkenmaier2023valitex}.

The most closely related work to our paper is the use of item response theory (IRT) models in NLP for constructing more informative test datasets~\cite{lalor2016building}, comparison of existing evaluation datasets and instances (e.g., difficulty, discrimination)~\cite{sedoc2020item,vania2021comparing,rodriguez2021evaluation,lalor2018understanding,rodriguez2022clustering}, as well as identification of difficult instances from training dynamics~\cite{lalor2020dynamic,lalor2019learning}.
Our work distinguishes itself from these papers in two aspects. First, we do not apply IRT to \textit{existing} LLM datasets/benchmarks. Instead, we introduce a framework for benchmarking LLMs by leveraging both IRT and test development knowledge from psychometrics. The goal of this framework is to generate new, high-quality benchmarks for LLMs that warrant valid comparison with human populations. 
Second, we demonstrate our framework with a mathematics proficiency test validated on 56 human populations, and compare LLM performance with human performance. To the best our knowledge, we are the first to apply psychometrically validated (mathematics) proficiency tests to LLMs and make valid model versus human comparison.

\section{Preliminaries}
In this section, we provide background knowledge on item response theory and test development in psychometrics. 
\subsection{Item Response Theory}
\label{sec:irt}
Item response theory (IRT) refers to a family of mathematical models that describe the functional relationship between responses to a test item, the test item's characteristics (e.g., item difficulty and discrimination) 
and test taker's standing on the latent construct being measured (e.g., academic proficiency)~\cite{standards2014}. Unlike classical test theory and current LLM benchmarks, which focus on the total or mean score of a test, IRT models takes into account the characteristics of both the items and the individuals (and models) being assessed, offering advantages like item quality diagnostics and more accurate estimation of test takers' proficiency. As such, IRT models have gained widespread adoption in various fields, including education, psychology, and healthcare, where trustworthy measurement and assessment are crucial.

We describe below three fundamental IRT models suitable for different types of test items: the 3-parameter logistic (3PL) model for multiple choice items scored as either incorrect or correct, the 2-parameter logistic (2PL) model for open-ended response items scored as either incorrect or correct, as well as the generalised partial credit (GPC) model for open-ended response items scored as either incorrect, partially correct, or correct.

The 3PL model gives the probability that a test taker, whose proficiency is characterised by the latent variable $\theta$, will respond correctly to item $i$:


\begingroup
\small
\begin{align}
\label{eq:3pl}
    & P\left(x_i=1 \mid \theta, a_i, b_i, c_i\right) \notag \\ 
    & = c_i+\frac{1-c_i}{1+\exp \left(-1.7 \cdot a_i \cdot\left(\theta-b_i\right)\right)} \\ 
    & \equiv P_{i, 1}\left(\theta\right) \notag
\end{align}
\endgroup

where
$x_i$ is the scored response to item $i$ ($1$ if correct and $0$ if incorrect); 
$\theta$ is the proficiency of the test taker, where a higher value implies a greater probability of responding correctly;
$a_i$ is the slope parameter of item $i$, characterising its discrimination (i.e., how well the item can tell test takers with higher $\theta$ from those with lower $\theta$)\footnote{The number $1.7$ is a scaling parameter to preserve historical interpretation of parameter $a_i$ on the normal ogive scale~\cite{camilli1994teacher}. Also applies to 2PL and GPC models.};
$b_i$ is the location parameter of item $i$, characterising its difficulty;
$c_i$ is the lower asymptote parameter of item $i$, reflecting the chances of test takers with very low proficiency selecting the correct answer (i.e., guessing). 
Correspondingly, the probability of an incorrect response to item $i$ is:
$   P_{i, 0}=P\left(x_i=0 \mid \theta_k, a_i, b_i, c_i\right)=1-P_{i, 1}\left(\theta_k\right)$. 
The 2PL model has the same form as the 3PL model (Equation \ref{eq:3pl}), except that the $c_i$ parameter is fixed at zero (i.e., no guessing).

The GPC model~\cite{muraki1992generalized} gives the probability that a test taker with proficiency $\theta$ will have, for the $i^{\text {th}}$ item, a response $x_i$ that is scored in the $l^{\text {th}}$ of $m_i$ ordered score categories:


\begingroup
\small
\begin{equation}
\label{eq:gpc}
\begin{aligned}
& P\left(x_i=l \mid \theta, a_i, b_i, d_{i, 1}, \cdots, d_{i, m_i-1}\right) \\
& = \frac{\exp \left(\sum_{v=0}^l 1.7 \cdot a_i \cdot\left(\theta-b_i+d_{i, v}\right)\right)}{\sum_{g=0}^{m_i-1} \exp \left(\sum_{v=0}^g 1.7 \cdot a_i \cdot\left(\theta-b_i+d_{i, v}\right)\right)}  \\
&\equiv P_{i, l}\left(\theta\right)
\end{aligned}
\end{equation}
\endgroup

where
$m_i$ is the number of response score categories for item $i$;
$x_i$ is the response score of item $i$ between 0 and $m_i-1$ (e.g., $0$, $1$ and $2$, for incorrect, partially correct, and correct responses);
$\theta$, $a_i$,  $b_i$ have the same interpretations as in the 3PL and 2PL models;
$d_{i, 1}$ is the category $l$ threshold parameter.
Setting $d_{i, 0}=0$ and $\sum_{j=1}^{m_i-1} d_{i, j}=0$ resolves the indeterminacy of the model parameters. 

Assuming conditional independence, the joint probability of a particular response pattern $x$ across a set of $n$ items is given by:


\begingroup
\small
\begin{equation}
\label{eq:likelihood}
    P\left(x \mid \theta, \text { item parameters }\right)=\prod_{i=1}^n \prod_{l=0}^{m_i-1} P_{i, l}\left(\theta\right)^{u_{i, l}}
\end{equation}
\endgroup

where $P_{i, l}\left(\theta\right)$ is of the form specific to the type of item (i.e., 3PL, 2PL or GPC); $m_i$ equals 2 for dichotomously scored items and 3 for polytomously scored items; $u_{i, l}$ is an indicator defined as:

\begingroup
\small
\[
u_{i, l}=\left\{\begin{array}{l}1 \text { if response } x_i \text { is in category } l \\ 0 \text { otherwise }\end{array}\right. 
\]
\endgroup

This function can be viewed as a likelihood function to be maximised by the item parameters. With the estimated item parameters, $\theta$ can then be estimated via various algorithms~\cite{reise2014handbook}.
In this paper, we use maximum likelihood because it gives an unbiased estimate of $\theta$. 

\subsection{Test Development in Psychometrics}
\label{sec:test_development}

\begin{table*}[ht]
\small
\begin{center}
\begin{tabular}{l|l}
\toprule
\textbf{Psychometrics} & \textbf{LLM Benchmarking} \\
\midrule
1. Construct and test need specification.& 1. (Construct and) test need specification.\\ 
2. Overall planning. & 2. Overall planning. \\ 
3. Item development. & 3. Dataset development. \\
\quad a. Construct refinement. & \quad a. Existing item collection \textit{OR} \\
\quad b. Item generation. & \quad \quad -- Quality control.  \\
\quad c. Item review. & \quad b. Item creation and/or annotation.\\
\quad d. Piloting of items. & \quad \quad -- Instructions. \\
\quad e. Psychometric quality analysis. & \quad \quad -- (Pilot) study. \\
4. Test construction and specification. & \quad \quad -- Agreement analysis. \\
5. Implementation and testing. & \quad \quad -- Error analysis. \\
6. Psychometric quality analysis. & 4. Dataset construction. \\
7. Test scoring and norming. & 5. Model selection and evaluation.   \\
8. Technical Manual. & 6. Benchmark release.   \\
\bottomrule
\end{tabular}
\end{center}
\caption{\textbf{Contrasting test development between psychometrics and LLM benchmarking.}}
\label{tab:test_development}
\end{table*}

Test development in psychometrics concerns the process of developing and implementing a test according to psychometric principles~\cite{irwing_test_2018}. Table~\ref{tab:test_development} contrasts psychometric test development (based on~\citet{irwing_test_2018}) with common LLM benchmarking procedures (based on~\cite{bowman2021will,raji2021ai}). In this section, we focus on the left panel -- psychometric test development. 

What sets psychometric test development apart from typical LLM benchmark development is its focus on ensuring that the test matches a well-defined construct via expert-driven item generation, rigorous pilot testing, use of factor analysis and IRT models for item and test diagnostics, establishment of scoring and normalisation standards, and testing on representative samples of intended test takers. 
The result of this elaborate process is a high-quality test that can assess the construct of interest for the test takers in a valid and reliable way. Many large-scale assessments, such as PISA (Programme for International Student Assessment), TIMSS and PIRLS (Progress in International Reading Literacy Study), conform to such a process. 

To further illustrate this process, we propose to use the example of assessing proficiency in grade school mathematics, which is a common construct of interest in psychometric testing and LLM benchmarking. For convenience, we abbreviate this construct as PGSM.

In Step 1, the construct of interest and the test need are specified. We ask, for instance, how do we define PGSM? Is it based on a specific curriculum? What does existing literature say? Which education levels are we interested in? Is the test meant for comparison between students within a school, or between schools within a country? Such questions help us to clarify what we want to measure and how it can be measured. 

In Step 2, we make necessary planning: How many test items? What kind of item format (e.g., multiple choice, short answer questions)? Will the test scores be standardised? How to assess the quality of test items? What are the desired psychometric properties of the test items (e.g., how discriminative and difficult should the items be?) and the test as a whole (e.g., internal consistency)? Will we pilot any test item? Will the test be computer- or paper-based? To sample test takers, what kind of sampling frames and strategies should we use? 

In Step 3, we develop test items, which is an iterative procedure involving five steps: (a) construct refinement, where we further clarify the definition of PGSM (e.g., What content domains should be included: number, algebra, and/or probability theory? Is proficiency only about knowing, or also about applying and reasoning?); (b) generate a pool of items with domain experts; (c) review the items for obvious misfit, errors and biases; (d) pilot the items with a representative sample of target test takers; (e) with the responses from the pilot step, we can assess the psychometric properties of the test items with IRT and factor analysis (e.g., item discrimination; item difficulty; factor structure\footnote{Factor structure refers to the correlational relationships between test items used to measure a construct of interest.}). 
We iterate this procedure until we have a set of test items with acceptable psychometric properties. Then, in Step 4, we construct the PGSM test by specifying, for instance, which items to include (if not all), in which order, how many equivalent test versions, and what scoring instructions to use.

In Step 5, the test gets implemented to the intended test takers, followed by Step 6: another round of quality analysis. If any item displays low quality characteristics (e.g., zero or negative discrimination), it will be left out of the final scoring.
In Step 7, responses of the test takers are scored for each item, and the resulting item-level scores form the basis for estimating proficiency scores using IRT or simpler procedures like (weighted) sums. It is typical to also normalise the proficiency scores (e.g., with a mean of 500 and a standard deviation of 100) to facilitate interpretations and comparisons.
Finally, in Step 8, a technical manual is compiled, detailing Step 1--7 and corresponding results, to facilitate correct re-use of the response data, the test, as well as interpretation of test scores, among other purposes.

\subsection{LLM Benchmark Development}
In this section, we focus on the right panel of Table~\ref{tab:test_development}: the process of LLM benchmark development, contrast it with test development in psychometrics and thereby highlight the potential benefits a psychometrics-based approach can introduction to LLM benchmarks. 

The process of developing LLM benchmarks is similar to test development in psychometrics. However, there are significant differences. To illustrate this, we take GSM8K~\cite{Cobbe2021TrainingVT} as an example, where we try our best to recreate the process of developing GSM8k based on the published dataset paper and map the specific steps to the six steps described in the right panel of Table~\ref{tab:test_development}.

First, the authors of GSM8K likely started by specifying the need for a large, high quality mathematics test at grade school level and of moderate difficulty for LLMs (Step 1). However, they did not explictly link the construct (i.e., PGSM) to any specific curriculum. 
Then, the authors made overall planning for the benchmark development (Step 2). For example, the number of items should be in the thousands; the crowd workers would curate the benchmark items; the authors would use agreement and error analysis to investigate the quality of the dataset; GPT-3 will be used to benchmark the dataset and verify dataset difficulty. 

In Step 3, namely dataset development\footnote{Note that we use the term ``dataset development'' here, contrasting ``item development'' in psychometrics, because of LLM benchmarks' typical emphasis on large and multiple datasets rather than concrete test items.}, often one of the two strategies is used: \textit{either} collect items from existing datasets and other sources and compile them into a new dataset, \textit{or}, create own items from scratch (with annotations). The authors of GSM8K followed the latter approach, an iterative procedure consisting of four parts: creating instructions (and possibly a user interface) for item generation and/or annotation; conducting a (pilot) study to collect the items and/or annotations; check annotator agreement; and assessing errors associated with the items or annotations. This step is iterated until a sufficient number of items and datasets are reached while meeting desired quality standards (e.g., high annotator agreement, low error rate). In total, GSM8K includes 8,500 items with solutions, with identified annotator disagreements resolved and a less than 2\%  error rate. 

In Step 4, the GSM8K authors compiled the final dataset from the crowdsourced items with training, evaluation and testing partitions. 
In Step 5, the GSM8k authors evaluated selected LLMs (i.e., GPT-3) on the dataset.
Finally, in Step 6, the authors released the benchmark, which consists of the dataset as well as its documentation (a research paper) and benchmarking results.

\paragraph{Comparison with Psychometrics}
\label{sec:comparison_with_psychometrics}
While sharing similarity with test development in psychometrics, current benchmark development for LLMs falls short on four aspects.
First, the construct of interest is often under-specified, leading to a mismatch between the intended construct and what the dataset actually measures. Again, take GSM8K as an example: While the dataset is intended to measure proficiency in grade school mathematics, the target grade level(s) are unclear and it only focuses on one content domain (algebra), missing other relevant ones like geometry and data. This is likely the result of not using established mathematics curricula and domain experts to develop test items. 

Second, despite researchers' interest in comparing LLM performance with human test takers (e.g., the GSM8K paper claims that ``a bright middle school student should be able to solve every problem''), such comparisons usually cannot be made because the test has not been designed with humans in mind or validated on any representative samples of the test's target user populations. 

Third, besides agreement and error analysis, LLM benchmarks can benefit from psychometric analysis of test items, (i.e., checking item discrimination and difficulty, as well as the factor structure of the items). While this is not yet the norm, there have been promising attempts (see Section~\ref{sec:related_work}). 

Lastly, the released benchmark often does not contain sufficient details about the process of benchmark creation. For instance, the GSM8K paper does not report instructions for item generation and annotation, results of the pilot study, agreement statistics, or annotator characteristics, all of which are important for external researchers to independently verify the quality of the benchmark.

\section{PATCH: \textbf{P}sychometrics-\textbf{A}ssis\textbf{T}ed ben\textbf{CH}marking of LLMs}

Figure~\ref{fig:framework} illustrates PATCH, our conceptualisation of a Psychometrics-AssisTed framework for benCHmarking LLMs against human populations. Each box represents a different research artifact, while each arrow applies an action to a source artifact and produces a subsequent target artifact.

Under PATCH, the first step is for researchers to define the construct of interest (e.g., proficiency in 8th grade mathematics), the specific LLM under examination, as well as the reference human population for comparison (e.g., 8th graders in Germany). 
Then, researchers look for an existing validated\footnote{The term ``validated'' means that the test has been (pre)tested on a representative sample of the target population of (human) test takers and fulfils psychometric quality requirements, such as sufficiently many discriminative items well distributed across different difficulty levels, and showing high reliability (e.g., high internal consistency) and validity (e.g., the sizes and directions of the empirical correlations among test items match theoretical expectations). } psychometric test measuring the specified construct; alternatively, a test can be developed from scratch by following the procedures described in Section~\ref{sec:test_development}, which likely requires collaboration with experienced psychometricians. 

Next, researchers use the items from the validated psychometric test to construct prompts for the LLM under evaluation and sample responses from the LLM. 
Similarly, researchers administer the psychometric test to a representative sample of the reference human population, and collect their responses. These human responses are then used to train and validate the IRT model(s) that match the type(s) of items in the psychometric test.

The resulting IRT model(s) will be fitted to the responses from the LLM and the humans, and will subsequently estimate each test taker's latent proficiency score---whether human or LLM---along with uncertainty estimates. These final proficiency scores\footnote{These latent proficiency scores are typically standardised $z$-scores (i.e., mean of $0$ and standard deviation of $1$), which sometimes go through further normalisation (e.g., re-scaling to a mean of $500$ and a standard deviation of $100$).} enable valid comparison between the LLM and the reference human population. 

At the heart of PATCH lies the psychometric test, which not only provides the basis for accurate measurement of the construct (i.e., capability of interest) but also enables valid comparison between LLMs and human test takers. Unfortunately, developing such a test can be a long and expensive process; utilising existing tests can be a shortcut, which should satisfy three conditions: 1) clear human population reference; 2) test items available in the public domain; 3) human responses and/or item parameter estimates available. 
The second and third are in practice difficult to meet, as many test institutes do not make their test items public due to commercial interests (e.g., SAT) or the need to measure trends over time (e.g., PISA). Collaboration with test institutes would alleviate this problem and additionally mitigate the likely data contamination issue with many public benchmarks. 

To the best of our knowledge, among academic proficiency tests, only TIMSS and PIRLS tests from certain years can be readily used for PATCH-based LLM benchmarking without formal collaboration with test institutes. TIMSS measures proficiency in grade school mathematics and science (4th grade, 8th grade, and final year of secondary school), while PIRLS assesses reading comprehension in 9/10-year-olds. Both TIMSS and PIRLS are administered in a large number of geographical regions with representative student samples, enabling population-level comparisons. 
In the following section, we demonstrate PATCH by measuring several LLMs' proficiency in 8th grade mathematics, using the latest available data from TIMSS 2011. 

\section{Demonstration: Leveraging IRT in PATCH to Measure LLM Proficiency in 8th Grade Mathematics}
In this section, we conduct an experiment to demonstrate the benefits and differences IRT modelling can make under our PATCH framework, compared to the traditional benchmark scoring approach. Note that we focus on the IRT part of PATCH instead of also on test development, because we lack the resources to develop our own valid test, and that there is no existing LLM benchmark that has the same measurement target as TIMSS 2011 (i.e., 8th grade mathematics proficiency) to enable comparison in terms of test development. Nevertheless, we hope our detailed comparison between LLM benchmark devleopment and psychomtric test development in Section~\ref{sec:comparison_with_psychometrics} suffices to fill in this gap. 

\subsection{Data: TIMSS 2011 8th Grade Mathematics}
\label{TIMSS2011}
56 geographical regions participated in TIMSS 2011, with typically a random sample of about 150 schools in each region and a random sample of about 4,000 students from these schools. These sample sizes are determined on the basis of a $\leq .035$ standard error for each region's mean proficiency estimate.
The use of random sampling makes unbiased proficiency estimates possible at the population level.
TIMSS 2011 has released a publicly available database\footnote{\url{https://timssandpirls.bc.edu/timss2011/international-database.html}}, of which three components are relevant to our study:

\paragraph{Test Items}
The TIMSS 2011 study has released 88 mathematics test items, 48 of which are multiple choice, 30 open-ended items scored as either incorrect or correct, and 10 open-ended items scored as either incorrect, partially correct, or correct. These items assess four content domains representative of 8th grade mathematics curriculum (agreed upon by experts from participating regions): number, algebra, geometry, data and chance. Within each domain, items are designed to cover various subtopics (e.g., decimals, functions, patterns) and three cognitive domains: knowing, applying and reasoning. 
These test items are only available in a PDF file that can be downloaded from the NCES website, which includes also scoring instructions.\footnote{\url{https://nces.ed.gov/timss/pdf/TIMSS2011_G8_Math.pdf}} 
To extract them into a format compatible with LLMs, we used OCR tools to extract as much textual information as possible, converted mathematical objects (e.g., numbers, symbols, equations, tables) into LaTeX format (following earlier benchmarks like MATH)~\cite{hendrycks2021measuring} and figures into JPEG format. See Appendix~\ref{app:chap5:user_prompts} for examples. We have released this LLM-compatible version of test items, as well as an eighth grade science test dataset from TIMSS 2011, an advanced secondary school mathematics test dataset from TIMSS 2008, and an advanced secondary school physics test dataset from TIMSS 2008. See Appendix~\ref{app:TIMSS} for details.

\paragraph{IRT and Item Parameters}
The TIMSS 2011 database also specifies the IRT model used for each test item and contains the item parameter estimates (e.g., discrimination, difficulty), which we use to reconstruct the final IRT model for proficiency estimation and verification. 

\paragraph{Student Responses and Proficiency Estimates}
Lastly, responses of the sampled students to each test item and their proficiency estimates are also available, allowing us to construct proficiency score distributions for each region.

\subsection{LLMs: GPT-4, Gemini-Pro and Qwen with Vision Capability}
Considering that more than 1/3 of the test items contain visual elements, we selected four competitive vision language models: GPT-4 with Vision (GPT-4V), Gemini-Pro-Vision, as well as the open-source Qwen-VL-Plus and Qwen-VL-Max~\cite{bai2023qwen}. There are more LLMs with vision capability. However, our goal is to showcase PATCH, not to benchmark as many LLMs as possible.

A major concern in using these LLMs is data contamination, which is difficulty to check due to inaccessible (information about) training data. However, as our focus is on demonstrating the PATCH framework, data contamination is less worrying. Furthermore, data contamination is still unlikely for four reasons.
First, these test items are copyrighted, forbidding commercial use.
Second, the test items are hard to extract from the source PDF.
Third, to the best of our knowledge, these test items do not exist in current LLM mathematics benchmarks. 
Fourth, we prompted the selected LLMs to explain or provide solutions to the test items' IDs (available in the source PDF). All failed to recognise these specific test IDs. 

\subsection{Prompts and Temperature}
We design two separate prompts for each test item: the system message and the user message.
We design the system message according to the prompt engineering guide by OpenAI, utilising chain-of-thought and step-by-step instructions on how to respond to the user message (i.e., with a classification of question type, an explanation and an answer (key)).\footnote{\url{https://platform.openai.com/docs/guides/prompt-engineering}} The system message is the same for all test items (see Appendix~\ref{app:chap5:system_prompt}).
Furthermore, to account for LLMs' sensitivity to slight variations in prompts~\cite{sclar2023quantifying,loya2023exploring}, we generate 10 additional variants of the system prompt with slight perturbations (e.g., lowercase a heading, vary the order of unordered bullet points). 

The user message is item-specific, containing both the item's textual description and the associated image(s) in base 64 encoded format.
See Appendix~\ref{app:chap5:user_prompts} for examples.\footnote{We are aware of other prompt engineering techniques like few-shot prompting and self-consistency. We did not experiment with them, as our focus is on demonstrating PATCH.}

Following~\citet{achiam2023gpt}'s technical report, we set the temperature parameter at 0.3 for multiple choice items and 0.6 for the others. See Appendix~\ref{app:chap5:responses} for example responses.

\begin{figure*}[ht]
    \centering
    \includegraphics[width=\linewidth]{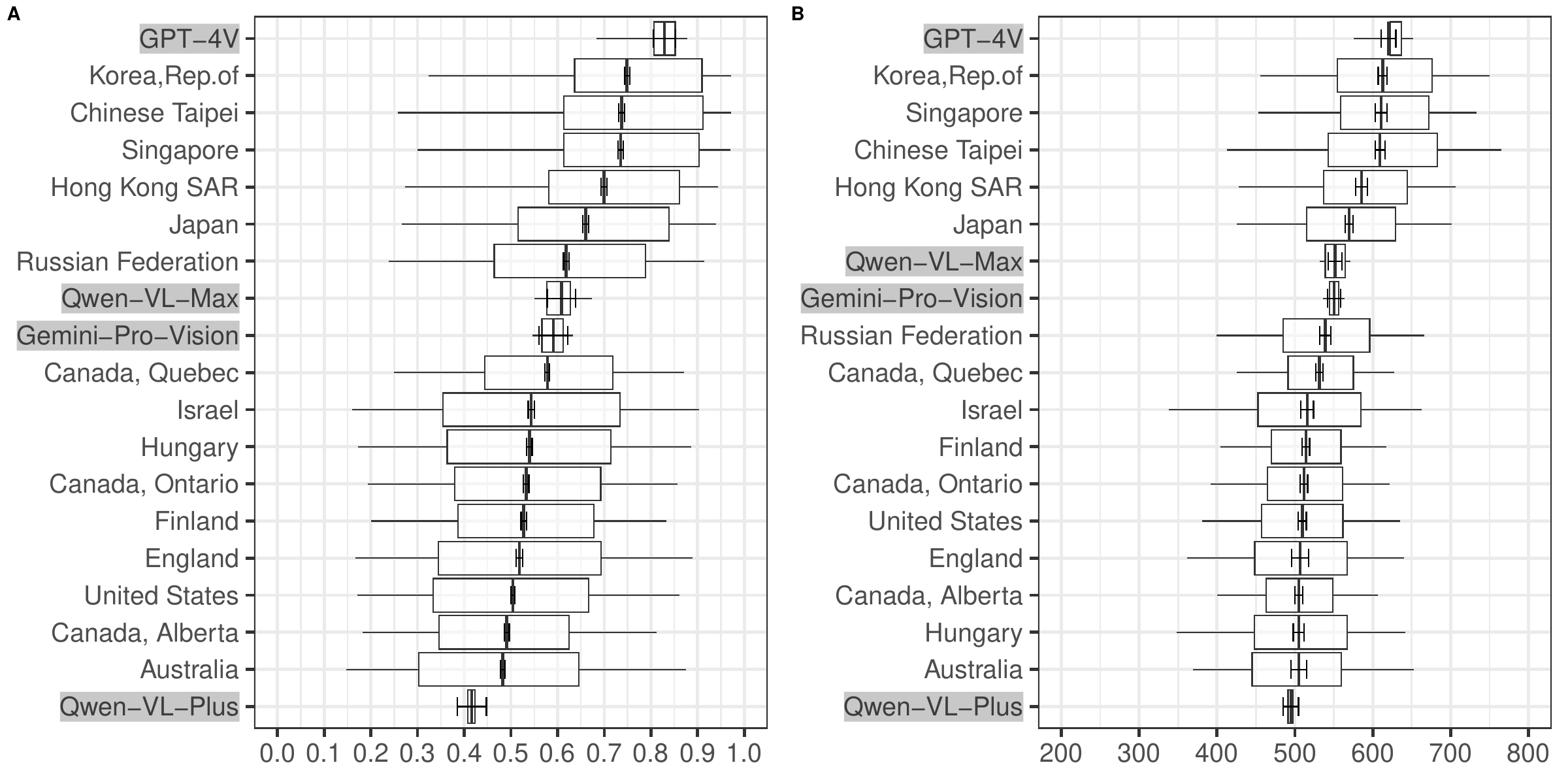}
    \caption{\textbf{Distribution of proficiency estimates for GPT-4V, Gemini-Vision-Pro, Qwen-VL-Plus, Qwen-VL-Max and selected participating regions of the TIMSS 2011 8th grade mathematics test.} Left figure (A) shows the proficiency estimates based on the percentages of correct responses. Right figure (B) shows the IRT-based proficiency estimates. The middle vertical line in each box plot represents the weighted mean proficiency score, with the error bars indicating its 95\% confidence interval. The borders of each box indicate the range of the middle 50\% of all values, with the two whiskers indicating the 5th and 95th percentiles. \textit{Note that we adhere to the official naming conventions of TIMSS 2011 when reporting the names of participating regions, with no intent to offend anyone}.}
    \label{fig:results}
\end{figure*}

\subsection{Scoring and Proficiency Estimation}
We manually scored the sampled responses from the LLMs following the official scoring rubrics of TIMSS 2011. 
Then, for multiple choice items, we apply the 3PL model (Equation~\ref{eq:3pl}); for open-ended items, we apply the GPC model (Equation~\ref{eq:gpc}) if partially correct response is admissible, otherwise the 2PL model.
We use maximum likelihood to obtain unbiased estimates of model proficiency scores ($\theta$) with the \texttt{mirt} package in R~\cite{chalmers2012mirt}.
This results in 11 $\theta$ estimates per model corresponding to 11 system message variants. We then use inverse variance weighting~\cite{marin2010weighting} to combine these estimates. Inverse variance weighting gives more weight to estimates that are more precise (i.e., having lower variance) and less weight to those that are less precise (i.e., having higher variance). This way, we obtain a more accurate \textit{overall} $\theta$ estimate and its 95\% confidence interval (CI) for each model. This further allows us to visually assess statistical significance by checking for overlap between CIs: for two independent samples, non-overlapping intervals suggest significance at $\alpha = 0.01$; slight overlap may still imply significance at $\alpha = 0.05$; and substantial overlap indicates non-significance at $\alpha = 0.05$.

\subsection{Results}
Figure~\ref{fig:results} shows the proficiency score distribution and ranking of the top 15 performing participating regions, as well as GPT-4V, Gemini-Pro-Vision, Qwen-VL-Plus and Qwen-VL-Max. The complete figures can be found in Appendix~\ref{app:chap5:results}.
The proficiency scores (x-axis) on the left panel are percentages of correct responses, corresponding to the default approach in current LLM benchmarking; the proficiency estimates on the right panel are based on IRT. We make four observations:

First, regardless of the method of proficiency estimation, the proficiency estimates show that GPT-4V has the overall best performance relative to the other models and 8th grade students of each participating region. 

Second, still looking at only the proficiency estimates, the method of proficiency estimation affects the ranking results. For instance, while Chinese Taipei is ranked 3rd on the left, it is ranked 4th on the right; Gemini-Pro-Vision is ranked 8th on the left, but ranked 7th on the right. Similarly, while Hungary is ranked 11th on the left, it drops to the 16th place on the right. 

Based on the first and second observations, one might argue that the overall rankings are similar (despite some large deviations such as Hungary) and therefore, IRT does not make a real difference in benchmarking LLMs and human populations. However, the similarity between the results of these two estimation approaches is to be expected, as the TIMSS items are already validated and thus of high-quality (i.e., good measurement properties, of various difficulty levels, high discrimination), rendering the use of IRT-based proficiency estimates not necessarily substantially different from using simple aggregate scores. Had the items been inconsistent in terms of measurement quality, more notable differences would have been observed.
Furthermore, our next observation, which focuses on uncertainty estimates, which are necessary on top of proficiency estimates for making claims about performance difference, lends support to our claimed importance of IRT. 

Third, the method of proficiency estimation affects the estimated 95\% CIs of human populations, which are usually wider when IRT is used. Notably, while on the left panel the CI of GPT-4V does not overlap with the second best (Rep. of Korea), indicating a statistically significant difference (\textit{t}(10.26) = 3.19, \textit{p} $<$ 0.01), they overlap on the right panel, suggesting otherwise (\textit{t}(10.40) = 1.25, \textit{p} = .24). This means that based on traditional proficiency estimation, GPT-4V performs significantly better than all the human groups, suggesting super-human performance. In contrast, when IRT is used, GPT-4V shares the same rank with Rep. of Korea, Singapore and Chinese Taipei, rejecting super-human performance.

Fourth, on the right panel, the CIs of the LLMs' proficiency estimates are generally narrower than their counterparts on the left panel, indicating that using IRT leads to more precise proficiency estimates for LLMs, a further advantage of IRT.

These findings show that the adoption of IRT with PATCH is likely to make a difference to LLM benchmark results, especially in contrast with human performances.

\section{Conclusion}
In this paper, we propose PATCH, a psychometrics-inspired framework to address current limitations of LLM benchmarks, including questionable measurement quality, lack of quality assessment on the item level and unwarranted comparison between humans and LLMs. 
We demonstrate the IRT part of PATCH with an 8th grade mathematics proficiency test and show evaluation outcomes that diverge from those based on existing benchmarking practices, especially when comparison with human test takers is made. This underscores the potential of PATCH to reshape the LLM benchmarking landscape.
Furthermore, we release 4 datasets that meet the requirements of PATCH, supporting the measurement of LLM proficiency in grade school math and science and its comparison with human performance. 
We hope to bring the LLM research community a step forward towards more scientific benchmarking and inspire more research in this direction. We also encourage researchers to collaborate with test institutes when developing new benchmarks, especially those that aim to measure cognitive capabilities. 


\section*{Limitations}
Our paper has the following limitations, among others. First,  PATCH requires validated tests, which can be resource-intensive if tests need to be developed from scratch. However, this also opens up opportunities for collaboration between LLM researchers, psychometricians and test institutes. 
Second, the validity, reliability, and fairness of using tests validated solely on humans for LLM benchmarking are debatable due to possibly differing notions of proficiency and cognitive processes between LLMs and humans. Nonetheless, such tests are still better than non-validated benchmarks, particularly for comparison of model and human performance.
Advancing LLM benchmarking further requires tests validated on LLMs (and humans for model-human comparisons), necessitating theoretical work on LLM-specific constructs and the development of LLM-specific IRT models and testing procedures.
Third, our experiment only includes four LLMs and one proficiency test. We consider this sufficient for demonstrating PATCH, but not enough if the goal is to benchmark as many LLMs as possible across different tests.

\section*{Acknowledgements}
We thank our reviewers, Anna Wegmann, Yupei Du, Melody Sepahpour-Fard, Elise Herrewijnen, Gianluca Sperduti for their helpful suggestions and comments. This work was supported by the Dutch Research Council (NWO) (grant number VI.Vidi.195.152 to D. L. Oberski; grant number VI.Veni.192.130 to D. Nguyen).

\bibliography{custom}

\appendix
\section{Prompts}

\subsection{Example Test Items (User Messages)}
\label{app:chap5:user_prompts}

\paragraph{Example 1}
\begin{quote}
The fractions $\frac{4}{14}$ and $\frac{\square}{21}$ are equivalent. What is the value of $\square$ ?

[A] 6 \hspace{2pt} 
[B] 7 \hspace{2pt} 
[C] 11 \hspace{2pt}
[D] 14
\end{quote}

\paragraph{Example 2}

\begin{quote}
\includegraphics[width=\linewidth]{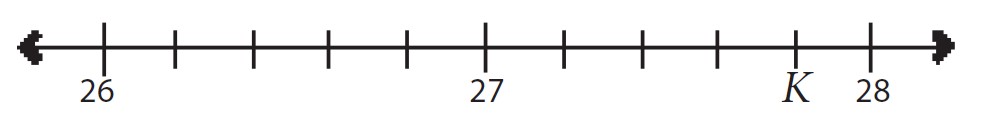}

Which number does $K$ represent on this number line?

[A] 27.4 \hspace{2pt} 
[B] 27.8 \hspace{2pt} 
[C] 27.9 \hspace{2pt} 
[D] 28.2

\end{quote}

\paragraph{Example 3}

\begin{quote}
\includegraphics[width=\linewidth]{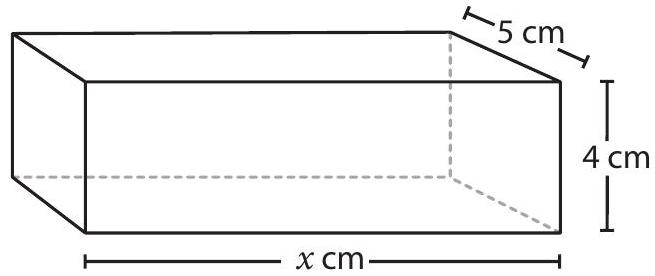}

The volume of the rectangular box is $200 \mathrm{~cm}^{3}$. What is the value of $x$ ?
\end{quote}

\subsection{Example System Messages}
\label{app:chap5:system_prompt}

Base prompt:
\begin{quote}
You are given a mathematics question written in LaTeX format.

Instructions: 

1. Type of question: Is it multiple choice, free text response, or drawing?

2. Think step by step, and describe your thought process and reasoning.

3. Answer: 

- For multiple choice: [selected answer key].

- For free-text response: [provide your short answer].

- For drawing: [describe clearly the steps to complete the drawing].

- If uncertain, make an educated guess.
\end{quote}

Variant 1 (item type reordered):
\begin{quote}
You are given a mathematics question written in LaTeX format.

Instructions: 

1. Type of question: Is it drawing, free text response, or multiple choice?

2. Think step by step, and describe your thought process and reasoning.

3. Answer: 

- For drawing: [describe clearly the steps to complete the drawing].

- For free-text response: [provide your short answer].

- For multiple choice: [selected answer key].

- If uncertain, make an educated guess.
\end{quote}

Variant 2 (new lines removed):
\begin{quote}
You are given a mathematics question written in LaTeX format.
Instructions: 
1. Type of question: Is it multiple choice, free text response, or drawing?
2. Think step by step, and describe your thought process and reasoning.
3. Answer: 
- For multiple choice: [selected answer key].
- For free-text response: [provide your short answer].
- For drawing: [describe clearly the steps to complete the drawing].
- If uncertain, make an educated guess.
\end{quote}

Variant 3 (upper case heading):
\begin{quote}
You are given a mathematics question written in LaTeX format.

INSTRUCTIONS: 

1. Type of question: Is it multiple choice, free text response, or drawing?

2. Think step by step, and describe your thought process and reasoning.

3. Answer: 

- For multiple choice: [selected answer key].

- For free-text response: [provide your short answer].

- For drawing: [describe clearly the steps to complete the drawing].

- If uncertain, make an educated guess.
\end{quote}

\section{Example Responses}
\label{app:chap5:responses}

Note that the example responses below correspond to (the order of) the example test items in Appendix~\ref{app:chap5:user_prompts}.

\paragraph{Example 1 (base prompt, GPT-4-V, correct)}
\begin{quote}
Type of question: Multiple choice

To find the value of the square that makes the fractions equivalent, we need to find a number that, when divided by 21, gives the same value as 4 divided by 14. 

To simplify the fraction $\frac{4}{14}$, we divide both the numerator and the denominator by their greatest common divisor, which is 2. This gives us $\frac{2}{7}$. 

Now, we need to find a number that, when placed in the square and divided by 21, gives us the fraction $\frac{2}{7}$. 

Since $21 = 7 \times 3$, we need to find a number that, when multiplied by $\frac{1}{3}$, gives us 2. That number is $2 \times 3 = 6$. 

Therefore, the value of the square that makes the fractions equivalent is 6.

Answer: [A] 6

Answer:
- For multiple-choice: [A] 6
\end{quote}

\paragraph{Example 2 (base prompt, Gemini-Vision-Pro, incorrect)}
\begin{quote}
 1. This is a multiple-choice question.
 
2. First, I notice that the number line is labelled from 26 to 28. This means that the value of $K$ must be between 26 and 28.

3. Next, I look at the answer choices. [A] 27.4 is the closest to 27, so it is a possibility. [B] 27.8 is also close, but it is further away from 27 than [A] 27.4. [C] 27.9 is even further away from 27 than [B] 27.8, so it is unlikely to be the correct answer. [D] 28.2 is the furthest away from 27, so it is the least likely to be the correct answer.

4. Based on this reasoning, I would select [A] 27.4 as the best answer.
\end{quote}

\paragraph{Example 3 (prompt variant 1, Qwen-VL-Max, correct)}
\begin{quote}








Is it drawing, free text response, or multiple choice? It's a multiple-choice question.

To find the volume of a rectangular prism (box), we use the formula:

$$\text{Volume} = \text{Length} \times \text{Width} \times \text{Height}$$

In this case, the length is $x$, the width is $5 \text{ cm}$, and the height is $4 \text{ cm}$. So,

$$200 \text{ } \text{cm}^3 = x \times 5 \text{ cm} \times 4 \text{ cm}$$

Now, let's solve for $x$:

$$x = \frac{200 \text{ } \text{cm}^3}{5 \text{ cm} \times 4 \text{ cm}}$$

$$x = \frac{200 \text{ } \text{cm}^3}{20 \text{ cm}^2}$$

$$x = 10 \text{ cm}$$

So, the value of $x$ is $10 \text{ cm}$. Answer Key: A) $10 \text{ cm}$

\end{quote}

\newpage

\section{TIMSS Datasets}
\label{app:TIMSS}
\paragraph{TIMSS 2011 Mathematics Eighth Grade}
This dataset was used in this study to demonstrate the PATCH framework. See Section~\ref{TIMSS2011} for details.

\begin{quote}
    Source: User Guide, Items and International Database for TIMSS 2011: Science -- Eighth Grade. Copyright \textcopyright 2013 International Association for the Evaluation of Educational Achievement (IEA). Publisher: TIMSS \& PIRLS International Study Center, Lynch School of Education, Boston College.
\end{quote}

Our study contributes three additional datasets. Similar to the dataset above, they are also based on officially released items by TIMSS but differ in the test subject, school grade level and/or test year. We constructed each dataset by using a mix of manual labour and OCR tools to extract item details from the official PDFs of the released items. The resulting dataset
consists of a LaTeX file ("main.tex") and a folder of item-related images. The test items are formatted in LLM-friendly format. With these three additional datasets, we hope to facilitate interested researchers to benchmark LLMs using these datasets with our PATCH framework. See below for more detail.

\paragraph{TIMSS 2011 Mathematics Fourth Grade}
This dataset is similar to the one we used to demonstrate PATCH but focuses on a different fourth grade mathematics with 73 items covering three domains: number, geometric shape and measures, and data display. It can be used to benchmark LLMs against representative samples of fourth-grade students from 57 regions.

\begin{quote}
    Source: User Guide, Items and International Database for TIMSS 2011: Mathematics -- Fourth Grade. Copyright \textcopyright 2013 International Association for the Evaluation of Educational Achievement (IEA). Publisher: TIMSS \& PIRLS International Study Center, Lynch School of Education, Boston College.
\end{quote}

\paragraph{TIMSS 2008 Advanced Mathematics}
This dataset focuses on assessing proficiency in advanced mathematics at the end of secondary high school. It can be used to benchmark LLMs against representative samples of final-year students in secondary school from 10 countries who have taken an advanced mathematics course. There are 40 items in total, covering algebra, calculus and geometry. 

\begin{quote}
    Source: TIMSS Advanced 2008 User Guide and Items for the International Database: Advanced Mathematics. Copyright \textcopyright 2009 International Association for the Evaluation of Educational Achievement (IEA). Publisher: TIMSS \& PIRLS International Study Center, Lynch School of Education, Boston College.
\end{quote}

\paragraph{TIMSS 2008 Advanced Physics}
This dataset focuses on assessing proficiency in advanced physics at the end of secondary high school. It can be used to benchmark LLMs against representative samples of final-year students in secondary school from 10 countries who have taken an advanced physics course. There are 39 items in total, covering mechanics, atomic and nuclear physics, electricity and magnetism, as well as heat and temperature. 

\begin{quote}
    Source: TIMSS Advanced 2008 User Guide and Items for the International Database: Advanced Physics. Copyright \textcopyright 2009 International Association for the Evaluation of Educational Achievement (IEA). Publisher: TIMSS \& PIRLS International Study Center, Lynch School of Education, Boston College.
\end{quote}

\paragraph{Licences}
According to the website of TIMSS 2011\footnote{\url{https://timssandpirls.bc.edu/timss2011/international-database.html}} and 2008\footnote{\url{https://timssandpirls.bc.edu/timss_advanced/idb.html}}:
\begin{quote}

TIMSS and PIRLS are registered trademarks of IEA. Use of these trademarks without permission of IEA by others may constitute trademark infringement. Furthermore, the website and its contents, together with all online and/or printed publications and released items by TIMSS, PIRLS, and IEA are and will remain the copyright of IEA.
    
All publications and released items by TIMSS, PIRLS, and IEA, as well as translations thereof, are for non-commercial, educational, and research purposes only. Prior notice is required when using IEA data sources or datasets for assessments or learning materials. IEA reserves the right to refuse copy deemed inappropriate or not properly sourced.

\end{quote}

Therefore, our use of TIMSS data in this research is in accordance with the intended use. 

\section{Use of AI Assistants}
We used ChatGPT to improve the writing of limited parts of the paper. We also used Mathpix to perform OCR on the PDFs containing the TIMSS released items before further processing into appropriate format. No AI was used for coding or analyses.

\section{Detailed Result Figure}
\label{app:chap5:results}
See next page.

\begin{figure*}[ht]
    \centering
    \includegraphics[width=\linewidth]{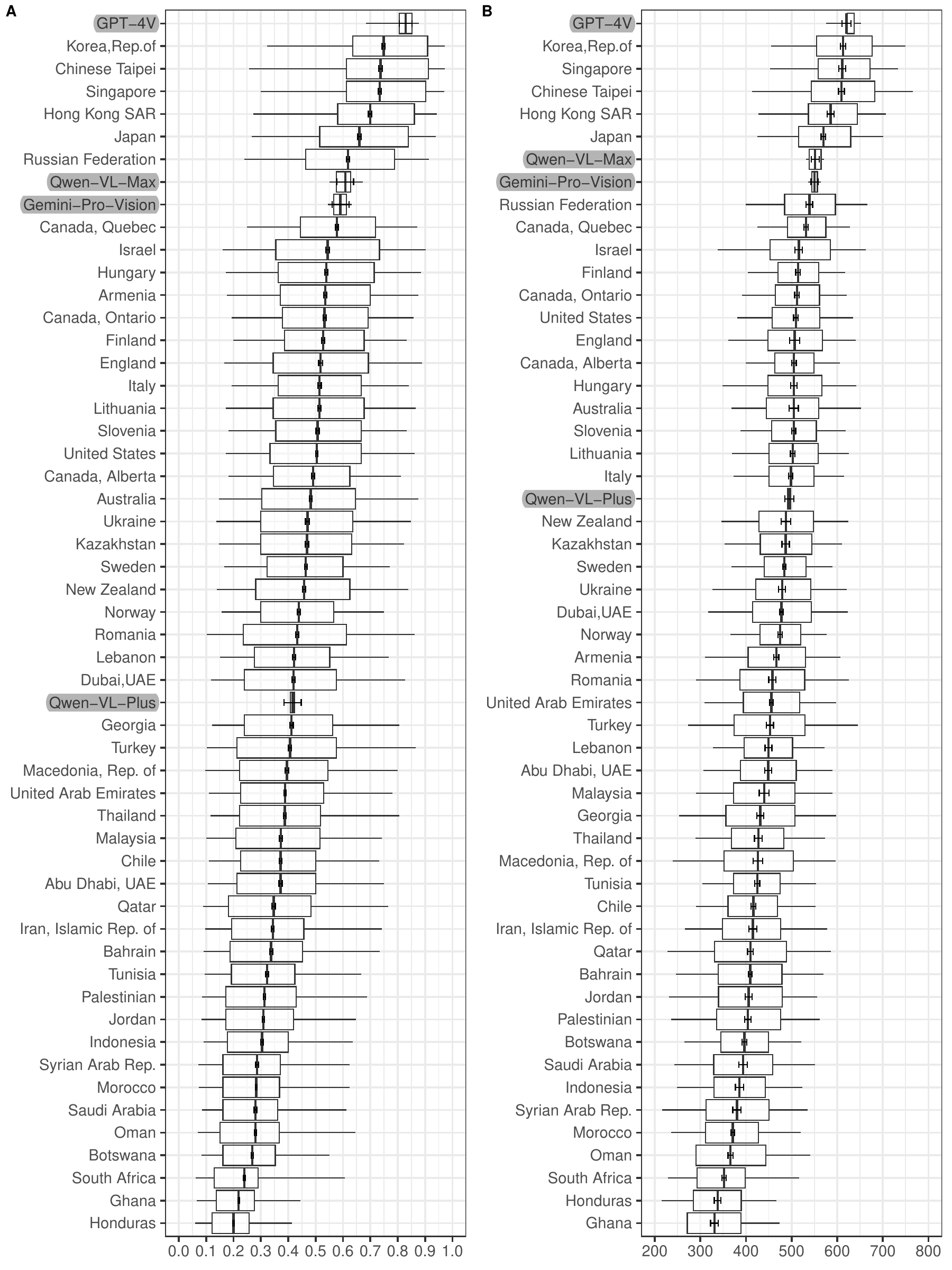}
    \caption{\textbf{Distribution of proficiency estimates for GPT-4V, Gemini-Vision-Pro, Qwen-VL-Plus, Qwen-VL-Max and all participating regions of TIMSS 2011 8th grade mathematics test.} Left figure (A) shows the proficiency estimates based on the percentages of correct responses. Right figure (B) shows the IRT-based proficiency estimates. The middle vertical line in each box plot represents the weighted mean proficiency score, with the error bars indicating its 95\% confidence interval. The borders of each box indicate the range of the middle 50\% of all values, with the two whiskers indicating the 5th and 95th percentiles. \textit{Note that we adhere to the official naming conventions of TIMSS 2011 when reporting the names of participating regions, with no intent to offend anyone}.}
    \label{fig:detailed_results}
\end{figure*}

\end{document}